\documentclass[letterpaper,journal]{IEEEtran}
\usepackage{amsmath,amsfonts}
\usepackage{array}
\usepackage[caption=false,font=normalsize,labelfont=sf,textfont=sf]{subfig}
\usepackage{textcomp}
\usepackage{stfloats}
\usepackage{url}
\usepackage{verbatim}
\usepackage{graphicx}
\usepackage{cite}
\usepackage{booktabs}
\begin{document}

\title{Localization-Infused Vision-Language Semantic Fusion for Text-Guided Medical Image Segmentation}

\author{Songyue Han, Mingye Zou, Shuchang Ye, Lei Bi,~\IEEEmembership{Member,~IEEE}, and Mingyuan Meng,~\IEEEmembership{Member,~IEEE}
\thanks{Songyue Han, Mingye Zou, and Mingyuan Meng are with the Zhongguancun Academy \& Zhongguancun Institute of Artificial Intelligence, Beijing, China; Mingye Zou is also with the Harbin Institute of Technology, Harbin, China. (e-mail: CooperLLM@126.com; s-zmy24@bza.edu.cn; mengmingyuan@bza.edu.cn).}
\thanks{Shuchang Ye is with the school of Computer Science, the University of Sydney, Sydney, NSW, Australia (e-mail: shuchang.ye@sydney.edu.au).}
\thanks{Lei Bi is with the Institute of Translational Medicine, Shanghai Jiao Tong University, Shanghai, China (e-mail: lei.bi@sjtu.edu.cn).}
\thanks{\textit{Preprint under review. Corresponding author: Mingyuan Meng.}}
}

\maketitle

\begin{abstract}
Medical image segmentation is essential for modern computer-aided medicine. Recently, text-guided segmentation has shown promise by incorporating clinician-formulated textual reports as semantic guidance for image segmentation. These textual reports contain language descriptions about the appearance, location, and neighboring anatomy of segmentation targets, providing explicit guidance for target localization and delineation. Existing text-guided segmentation methods typically extract textual semantics implicitly through a pretrained text encoder and then integrate vision-language semantics via straightforward image-text feature fusion. However, these methods do not explicitly capture target-oriented information embedded in textual reports, particularly target location, and do not explore multi-level information fusion strategies beyond basic feature-level fusion, limiting the extraction and integration of critical textual semantics. In this study, we propose LoG, a localization-infused vision-language fusion framework for text-guided medical image segmentation. By jointly performing multi-scale target localization tasks, LoG explicitly captures target-oriented vision-language semantics and enables three-level localization-infused semantic fusion: (i) localization-guided feature fusion that directly infuses location-relevant semantics into visual features, (ii) localization-gated attention fusion that redirects multi-scale localization predictions to reinforce critical regions, and (iii) localization-constrained loss fusion that supervises segmentation based on spatial consistency with target localization. Extensive experiments on three well-established benchmark datasets, involving three medical imaging modalities with paired textual reports, demonstrate that LoG consistently outperforms state-of-the-art medical image segmentation methods.
\end{abstract}


\section{Introduction}
\IEEEPARstart{M}{edical} image segmentation plays an essential role in
a wide range of modern medical applications, such as computer-aided diagnosis, surgical navigation, and quantitative assessment of disease progression \cite{Hesamian2019}. Its goal is to identify regions of interest (ROIs) in medical images. For example, identifying and quantifying tumor ROIs in patients with cancer is crucial for assessing disease progression and treatment effectiveness \cite{JiangXiaoyan2023}. Unfortunately, manual segmentation is labor-intensive and time-consuming, requires specialized expertise, and suffers from considerable inter-clinician variability. Therefore, extensive research has been devoted to automated segmentation, with deep learning driving substantial advances over the past decade \cite{Ronneberger2015, Zhou2018, oktay2018attention, Cao2023205, IsenseeFabian2020}. Despite these advances, most existing segmentation methods rely solely on image information and lack semantic and spatial priors regarding ROI identity and location. As a result, they are prone to missing subtle ROIs or yielding location drift when handling complex medical images with ambiguous boundaries and low-contrast anatomical structures.

To address these limitations, textual clinical reports have been incorporated as supplementary information for medical image segmentation, demonstrating the potential to surpass the performance bounds of conventional image-only segmentation \cite{ZhangJingyi2024}. These reports, formulated by clinical experts, are routinely available alongside medical images in clinical workflows, providing informative language descriptions about ROI appearance (e.g., shape, color, and opacity), location (e.g., upper-right, lower-left, etc.), and their contextual relationships with neighboring anatomical structures (e.g., near the right upper lobe). Early explorations of leveraging textual reports for image segmentation centered on vision-language pretraining (VLP) \cite{RadfordAlec2021}, where models pretrained on image-text data achieved improved performance when applied to downstream segmentation tasks \cite{HuangShih-Cheng2022, zhang2025multimodal}. However, the pretraining-based models are not optimized for segmentation tasks during pretraining and, once pretrained, cannot explicitly leverage textual information during inference. Recently, text-guided medical image segmentation methods have been developed to incorporate clinical reports as auxiliary input to guide ROI delineation, enabling superior segmentation performance over image-only and pretraining-based methods and setting new benchmarks in the field \cite{LiZihan2024, ZhongYi2023, LeeGo-Eun2024, ZengQingjie2026, BhardwajRahul2026, XueHengzhi2026,li2026language}.

From the perspective of information fusion, text-guided segmentation represents a cross-modal fusion problem in which complementary visual and textual information is integrated to achieve accurate pixel-wise ROI delineation. Its performance superiority over unimodal image-only methods largely hinges on two critical aspects in the extraction and integration of cross-modal textual semantics: (i) how to precisely extract key semantics about ROI identity and location from clinical reports and transform them into pixel-wise location cues favorable for segmentation tasks, and (ii) how to adequately fuse the extracted textual semantics with visual contexts to provide effective pixel-wise guidance on segmentation. Unfortunately, as illustrated in Fig.~\ref{fig1}(a), existing text-guided medical image segmentation methods, e.g., \cite{LiZihan2024, ZhongYi2023, LeeGo-Eun2024, ZengQingjie2026, BhardwajRahul2026, XueHengzhi2026, li2026language}, exhibit notable limitations in both aspects:

\begin{figure*}[t]
    \centering
    \includegraphics[width=\textwidth]{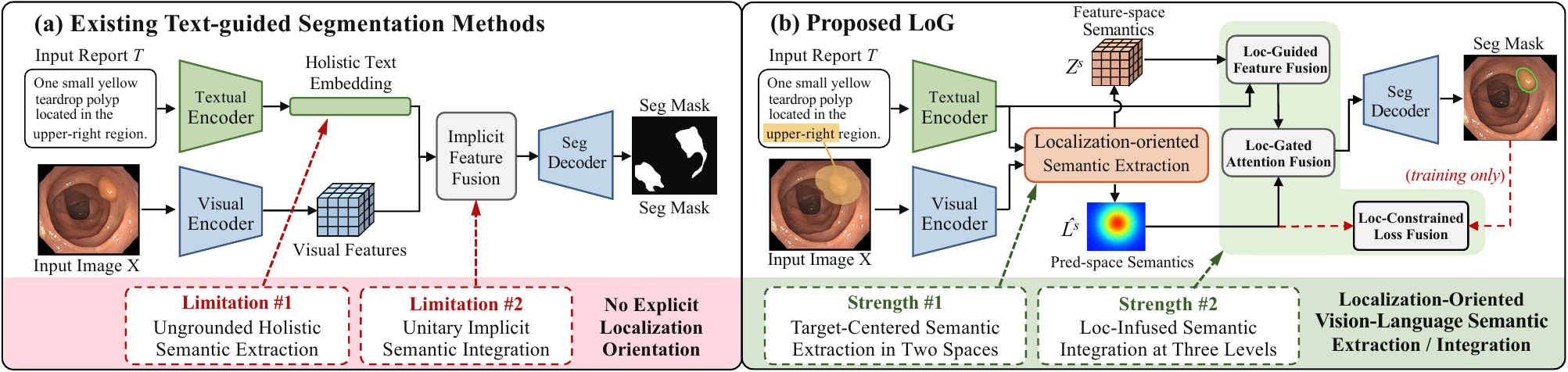}
    \caption{Illustration of existing text-guided medical image segmentation
    methods and the proposed LoG.
    (a) Existing methods typically encode the textual report into holistic
    text embeddings and fuse them with visual features through implicit
    feature-level fusion.
    (b) LoG explicitly extracts localization-oriented semantics from
    image-report inputs and injects them into segmentation via
    three-level localization-infused semantic fusion, including LGFF, LGAF,
    and LCLF.}
    \label{fig1}
\end{figure*}

First, existing methods typically extract textual semantics implicitly through a pretrained text encoder (e.g., CXR-BERT \cite{boecking2022textsemantics}), failing to explicitly capture target-oriented information, especially regarding ROI location, embedded in textual reports. They encode entire clinical reports into holistic textual features in which crucial semantics are diluted among redundant segmentation-irrelevant information, thereby struggling to extract reliable location cues to guide segmentation. Second, existing methods tend to rely on straightforward image-text feature fusion for integrating vision-language semantics while rarely exploring multi-level information fusion strategies for comprehensive utilization of cross-modal semantics. They predominantly focus on complicating feature-level fusion, developing from basic cross-attention in early works \cite{ZhongYi2023, LeeGo-Eun2024} to recent sophisticated strategies, e.g., dynamic convolution-based feature fusion \cite{BhardwajRahul2026} and progressive cross-scale feature alignment \cite{XueHengzhi2026}. However, other potentially complementary fusion strategies, such as gating-based modulation \cite{fu2021msam} or loss-level constraints \cite{bai2025tdfusion}, have rarely been attempted, and a holistic multi-level fusion strategy that collectively exploits cross-modal semantic guidance across multiple fusion levels remains unexplored as well.

To address these limitations, we propose a target-oriented multi-level fusion paradigm for integrating vision-language semantics, which collectively optimizes both the extraction and integration of cross-modal textual information. It explicitly captures target-relevant semantic cues, especially those concerning ROI identity and location, from clinical reports and then leverages these cues to guide a multi-level fusion process that integrates vision-language semantics at multiple granularities. Specifically, as illustrated in Fig.~\ref{fig1}(b), we instantiate this paradigm as LoG, a localization-infused vision-language fusion framework that reformulates text-guided segmentation as a localization-oriented semantic extraction and fusion problem. LoG jointly performs multi-scale ROI localization tasks to extract localization-oriented semantic cues from paired images and clinical reports. The extracted semantics are then embedded into an implicit feature space and an explicit localization-prediction space, enabling three-level localization-infused information fusion. First, Localization-Guided Feature Fusion (LGFF) infuses the localization-oriented semantics embedded in the feature space into the segmentation model, thereby providing implicit feature-level guidance that biases the model toward target regions during feature composition. The semantics embedded in the localization-prediction space, represented as multi-scale localization maps, are then redirected into the segmentation model to drive Localization-Gated Attention Fusion (LGAF) and Localization-Constrained Loss Fusion (LCLF). The former gates visual feature maps according to the predicted localization maps to modulate attention toward critical regions, while the latter exploits the predicted localization maps as multi-scale spatial consistency constraints on segmentation prediction.

Our main contributions can be summarized as follows:

\begin{itemize}
\item We propose a target-oriented multi-level fusion paradigm and instantiate it as LoG for text-guided medical image segmentation, reformulating both the extraction and integration of cross-modal textual semantics as localization-infused processes for more precise ROI delineation.
\item For textual semantic extraction, LoG innovatively incorporates multi-scale localization into text-guided segmentation, jointly capturing both explicit (localization maps) and implicit (feature embeddings) semantic cues that are directly oriented toward ROI identity and location.
\item For textual semantic integration, LoG introduces three-level localization-infused semantic fusion strategies, namely LGFF, LGAF, and LCLF, collectively injecting the extracted localization-oriented semantics into the segmentation model through comprehensive feature-level fusion, gating-based modulation, and loss-level constraints.
\end{itemize}

Extensive experiments on three well-established benchmark datasets (QaTa-COV19 \cite{degerli2022osegnet}, MosMedData+ \cite{morozov2020mosmeddata}, and Kvasir-SEG \cite{jha2020kvasirseg}), covering three medical imaging modalities (X-ray, CT, and colonoscopy) with paired textual reports, demonstrate that LoG consistently outperforms state-of-the-art medical image segmentation methods.

\section{Related Work}
\subsection{Text-guided Medical Image Segmentation}
Text-guided medical image segmentation aims to leverage textual clinical reports as auxiliary semantic guidance for ROI delineation. A pioneering work is LViT \cite{LiZihan2024}, which introduced paired image-report inputs into medical image segmentation and integrated visual and textual features through a U-shaped CNN and a Vision Transformer. Along with this work, Li et al. \cite{LiZihan2024} annotated the open-source segmentation datasets QaTa-COV19 \cite{degerli2022osegnet} and MosMedData+ \cite{morozov2020mosmeddata} with clinical reports, providing important benchmarks for subsequent studies. Following LViT, a series of text-guided segmentation methods have been developed to exploit textual semantics for improving segmentation accuracy \cite{LiZihan2024, ZhongYi2023, LeeGo-Eun2024, ZengQingjie2026, BhardwajRahul2026, XueHengzhi2026, li2026language, yu2025fmiseg, pan2026evivlm, huang2025reclmis}.

Most existing methods focus on designing cross-modal fusion modules to establish image-text semantic interactions. For example, Ariadne's Thread \cite{ZhongYi2023} integrated textual features with multi-scale visual features in the decoder. MAdaptor \cite{zhang2024madapter} proposed a bidirectional cross-modal fusion architecture to facilitate mutual information exchange between visual and textual features. FMISeg \cite{yu2025fmiseg} explored frequency-domain multimodal interaction by incorporating textual features into high- and low-frequency visual representations. TeViA \cite{ZengQingjie2026} introduced segmentation-specific text-to-vision alignment constraints to enhance textual feature integration. ViTexNet \cite{BhardwajRahul2026} further employed text-guided dynamic convolution to modulate visual features for segmentation. RecLMIS \cite{huang2025reclmis} proposed a cross-modal conditioned reconstruction framework to explicitly model image-text interactions through vision-language reconstruction. In addition, several studies have attempted to reduce reliance on textual input during inference, such as SGSeg \cite{ye2024sgseg} and ProLearn \cite{ye2025prolearn}, which enable text-assisted training with text-free inference.

Despite these advances, existing text-guided segmentation methods generally extract textual semantics in an implicit and holistic manner. They typically encode the entire report into global textual features; however, the key textual semantics may be diluted by redundant report content when represented as holistic feature embeddings. Further, most existing methods exclusively focus on feature-level image-text fusion. This motivates our method, which explicitly captures target-oriented semantics from paired images and reports and further injects them into segmentation through multi-level vision-language information fusion.

\subsection{Cross-modal Information Fusion}
Cross-modal information fusion aims to integrate complementary information from heterogeneous modalities to achieve comprehensive characterization. According to the stage at which multimodal information is integrated, existing fusion strategies can be broadly categorized into input-level, decision-level, and feature-level fusion. Input-level fusion directly combines raw inputs from different modalities, such as concatenating multi-sequence or multimodal medical images before feeding them into the model \cite{rallabandi2023mripet}. Decision-level fusion integrates the final predictions produced by different modality-specific models, commonly through voting, averaging, or weighted aggregation \cite{yang2023ensemble}. Feature-level fusion integrates modality-specific representations in intermediate network layers through concatenation, element-wise operations, or attention mechanisms \cite{zhu2025beam}. In particular, attention-based feature-level fusion has been extensively explored to model cross-modal semantic dependencies, with cross-attention becoming a dominant mechanism in text-guided segmentation \cite{ZhongYi2023, LeeGo-Eun2024}. Similar attention-based designs have also demonstrated effectiveness in broader multi-source fusion scenarios, such as AMEFuse \cite{liu2022amfuse} and ICAFusion \cite{shen2024icafusion}.

In addition, more advanced fusion strategies have been developed to enable adaptive and task-aware cross-modal interaction. One representative strategy is modulation/gating-based fusion, where information from one modality is used to modulate, gate, or filter the representation of another modality. For example, MSAM employs the prediction of one modality to spatially gate the features of another modality \cite{fu2021msam}, and modular gating structures are designed for stage-wise multimodal feature filtering \cite{zhang2024pregating}. In the medical domain, MsgFusion \cite{wen2024msgfusion} exploits medical semantic information to guide two-branch feature extraction for multimodal brain image fusion, indicating that high-level semantic cues can effectively steer cross-modal feature aggregation. Another representative strategy is loss-level fusion, where multimodal information interaction is imposed through optimization objectives rather than explicit feature aggregation. For example, MCFNet \cite{qiao2025mcfnet} dynamically balances multimodal and unimodal objectives, and DFusion \cite{bai2025tdfusion} introduces task-driven supervision to guide the generation of fusion parameters. Contrastive learning is also adopted to align cross-modal representations by enforcing semantic consistency between positive pairs \cite{RadfordAlec2021, chen2020simclr}. These studies suggest that effective information fusion can be achieved not only by directly combining features but also through adaptive modulation or loss-level optimization constraints.

For text-guided medical image segmentation, the adopted cross-modal fusion strategies remain dominated by feature-level image-text fusion. EviVLM \cite{pan2026evivlm} is an exception that introduces loss-level information interaction, but it still focuses mainly on cross-modal consistency and uncertainty modeling, without explicitly exploiting localization information to organize holistic multi-level fusion. Therefore, how to transform textual localization cues into segmentation-oriented guidance and systematically inject them across multiple fusion levels remains underexplored.

\section{Method}
Given a image $X\in\mathbb{R}^{C\times H\times W}$, where $C$ denotes the number of channels and $H$ and $W$ denote the spatial dimensions, and its associated report $T=\left\{w_{i}\right\}_{i=1}^{N}$ consisting of $N$ words, text-guided medical image segmentation aims to predict a pixel-wise segmentation mask $Y \in\{0,1\}^{H \times W}$. Conventional vision-only segmentation can be formulated as:
\begin{equation}
\label{eq1}
    \hat{Y}=D_\theta(E_v(X)),
\end{equation}where $E_{v}(\cdot)$ and $D_{\theta}(\cdot)$ denote a visual encoder and a segmentation decoder. Text-guided segmentation further incorporates textual information through a textual encoder $E_{t}(\cdot)$ and a cross-modal fusion operator $\mathcal{F}(\cdot)$, generally formulated as:

\begin{equation}
\label{eq2}
     \hat{Y}=D_\theta\left(\mathcal{F}(E_v(X),E_t(T))\right).
\end{equation}

\begin{figure*}[t]
    \centering
    \includegraphics[
        width=0.98\textwidth,
        keepaspectratio
    ]{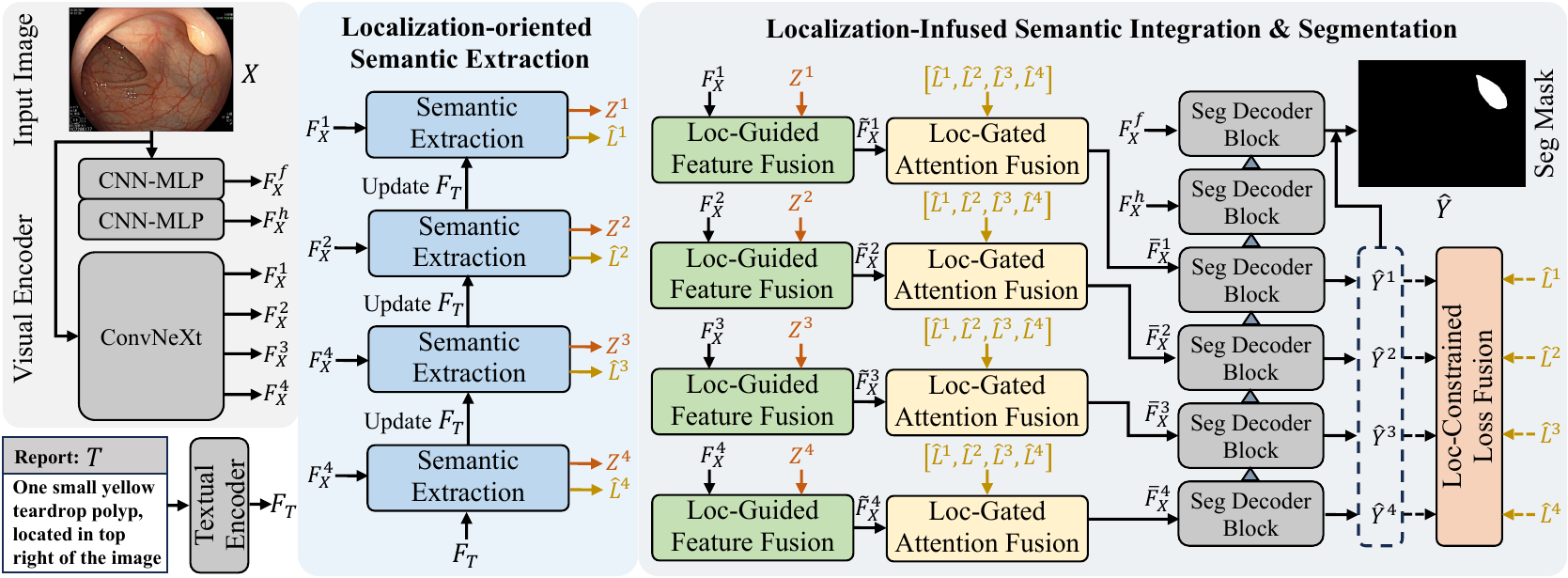}
    \caption{Overall workflow of the proposed LoG framework. LoG consists of
    a visual encoder, a textual encoder, and a localization-oriented
    vision--language semantic extraction process, followed by
    Localization-Guided Feature Fusion (LGFF),
    Localization-Gated Attention Fusion (LGAF),
    Localization-Constrained Loss Fusion (LCLF), and a segmentation decoder.}
    \label{fig2}
\end{figure*}

Fig.~\ref{fig2} illustrates the overall workflow of LoG, a localization-infused vision-language fusion framework for text-guided medical image segmentation. LoG takes the image-report pair $(X,T)$ as input and jointly predicts the final segmentation mask $\hat{Y}$ and a set of multi-scale localization maps $\left\{\hat{L}^{s}\right\}_{s=1}^{S}$ for $S$ hierarchical scales, with the latter serving as intermediate outputs for localization-infused fusion. Specifically, LoG first employs a visual encoder and a textual encoder to extract multi-scale visual features and token-level textual features (detailed in Section~\ref{sec:encoding}). Then, multi-scale localization prediction is performed using both visual and textual features to extract localization-oriented vision-language semantics and embed them in two spaces: an implicit feature space and an explicit localization-prediction space (detailed in Section~\ref{sec:extraction}). Based on these semantics, Localization-Guided Feature Fusion (LGFF, Section~\ref{sec:lgff}) and Localization-Gated Attention Fusion (LGAF, Section~\ref{sec:lgaf}) progressively inject localization-oriented semantic cues into visual features and reinforce target-relevant regions, followed by a segmentation decoder for final segmentation prediction. Finally, Localization-Constrained Loss Fusion (LCLF, Section~\ref{sec:lclf}) is defined as the overall training objective, which jointly optimizes segmentation prediction, multi-scale localization prediction, and localization-segmentation consistency.

\subsection{Visual and Textual Encoding}\label{sec:encoding}
For visual encoding, given an input image $X$, we employ ConvNeXt-Tiny \cite{liu2022convnext} as the visual encoder following Zhong et al. \cite{ZhongYi2023}. The visual encoder hierarchically extracts multi-scale features from four encoding stages of ConvNeXt-Tiny, denoted as $[F_X^1,F_X^2,F_X^3,F_X^4]$. These features correspond to spatial resolutions from 1/4 to 1/32 of the original image scale, respectively, providing both fine-grained spatial details from shallow layers and high-level semantic contexts from deep layers. To further preserve high-resolution image details, we follow \cite{meng2023fullresolution} and introduce two hybrid CNN-MLP stages to extract full-scale and half-scale visual features, denoted as $F_X^f$ and $F_X^h$. Each hybrid stage consists of parallel residual convolution and multi-axis gated MLP \cite{tu2022maxim} branches, where the convolution branch captures local structural details and the MLP branch models long-range visual dependencies. $F_X^f$ and $F_X^h$ are directly propagated via skip connections to the segmentation decoder to recover high-resolution details, without participating in cross-modal vision-language modeling.

For textual encoding, we adopt BiomedBERT \cite{gu2022pubmedbert} as the textual encoder, a medical domain-specific language model pretrained on large-scale biomedical literature. Given the associated textual report $T$, BiomedBERT produces contextualized token-level textual features that encode medical concepts and report semantics. We extract the hidden representations from the last layers of BiomedBERT, denoted as $F_{T}$, which are then used to interact with multi-scale visual features for localization-oriented semantic extraction.

\subsection{Localization-oriented Semantic Extraction}\label{sec:extraction}
To extract localization-oriented vision-language semantics, we perform multi-scale localization prediction based on the interaction between visual and textual features. Given the multi-scale visual features $[F_X^1,F_X^2,F_X^3,F_X^4]$ and textual feature $F_{T}$, multi-scale localization prediction is progressively conducted from deep to shallow stages across four scales.

\begin{figure}[t]
    \centering
    \includegraphics[
        width=0.48\textwidth,
        keepaspectratio
    ]{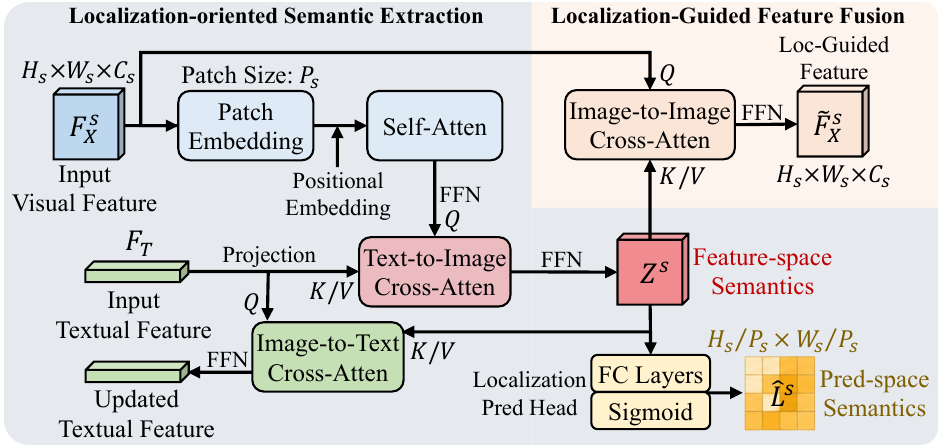}
    \caption{Internal structure of localization-oriented semantic extraction
    and Localization-Guided Feature Fusion (LGFF) at the scale $s$.}
    \label{fig3}
\end{figure}

For each scale $s\in\{4,3,2,1\}$, as illustrated in Fig.~\ref{fig3}, the visual feature $F_{X}^{s}$, with spatial dimensions $H_{s}\times W_{s}$, is first transformed into a sequence of non-overlapping image patch tokens through a patch embedding layer with patch size $P_{s}$. Positional embeddings are then added to retain spatial order. A self-attention block is applied to the image patch tokens to model intra-image contextual dependencies within the corresponding scale. Then, the textual feature is projected into the same embedding dimension as the image patch tokens and interacts with them through text-to-image cross-attention. Specifically, the image patch tokens serve as queries, while the projected textual tokens serve as keys and values. This cross-attention operation enables each image patch to selectively aggregate target-relevant textual semantics. The resulting tokens are further refined by a feed-forward network and denoted as $Z^{s}$, which represents the localization-oriented semantics in the feature space. Based on $Z^{s}$, a lightweight localization prediction head is used to generate the scale-specific localization prediction $\widehat{L}^{s}$, which represents the localization-oriented semantics in the prediction space. The localization prediction head consists of fully connected layers followed by a sigmoid activation, and the localization prediction is formulated as a patch-wise localization mask $\hat{L}^{s}\in[0,1]$ of size $H_s/P_s\times W_s/P_s$, where each element indicates the probability that the corresponding image patch contains the target ROI. After each scale, image-to-text cross-attention uses $Z^{s}$ to update the textual feature $F_T$; the updated textual feature is then passed to the next finer scale, allowing it to progressively absorb scale-specific visual contexts.

\subsection{Localization-Guided Feature Fusion (LGFF)}\label{sec:lgff}
Based on the feature-space localization semantics $Z^s$, Localization-Guided Feature Fusion (LGFF) is introduced to inject target-oriented vision-language information back into the visual feature pyramid. For each scale $s \in\{4,3,2,1\}$, as illustrated in Fig.~\ref{fig3}, the original visual feature $F_X^s$ is first flattened into image tokens and projected into the same hidden dimension as $Z^s$. Then, a cross-attention operation is performed, where the projected visual tokens are used as queries and the localization-oriented fused tokens $Z^s$ are used as keys and values. Through this design, each visual token adaptively retrieves target-relevant semantic cues from $Z^s$, enabling the visual tokens to be enhanced according to localization-aware image-text interactions. The resultant visual tokens are further processed by a feed-forward network and then projected back to the original channel dimension. Finally, the token sequence is reshaped into the visual feature map, yielding the localization-guided visual feature $\widetilde{F}_X^s$.

After applying LGFF at four scales, we obtain the localization-guided feature pyramid $\left[\widetilde{F}_X^1, \widetilde{F}_X^2, \widetilde{F}_X^3, \widetilde{F}_X^4\right]$, which is then passed to the subsequent LGAF and segmentation decoder. These processed features preserve the original multi-scale visual hierarchy while embedding target-relevant textual semantics, providing feature-level localization guidance for segmentation.

\subsection{Localization-Gated Attention Fusion (LGAF)}\label{sec:lgaf}
\begin{figure}[t]
    \centering
    \includegraphics[
        width=0.48\textwidth,
        keepaspectratio
    ]{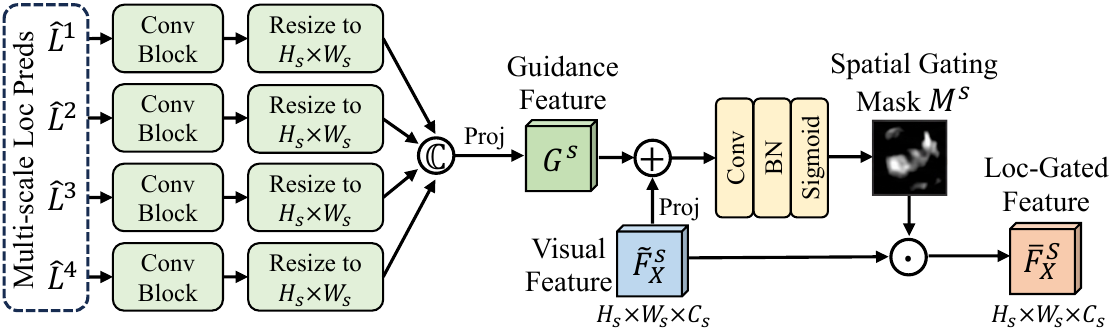}
    \caption{Internal structure of Localization-Gated Attention Fusion
    (LGAF) at the scale $s$.}
    \label{fig4}
\end{figure}

Based on the prediction-space localization semantics $\widehat{L}^{s}$, Localization-Gated Attention Fusion (LGAF) is introduced to further redirect the feature pyramid $\left[\widetilde{F}_X^1, \widetilde{F}_X^2, \widetilde{F}_X^3, \widetilde{F}_X^4\right]$ toward target-relevant regions. For each scale $s \in\{4,3,2,1\}$, LGAF takes the localization-guided visual feature $\widetilde{F}_X^s$ and all multi-scale localization predictions $\left[\hat{L}^1, \hat{L}^2, \hat{L}^3, \hat{L}^4\right]$ as input, and generates a spatial attention mask to reinforce potential ROI regions. Specifically, as illustrated in Fig.~\ref{fig4}, each localization map $\hat{L}^i$ is first processed by a convolutional block and all processed localization maps are then resized to the spatial resolution of $\widetilde{F}_X^s$ through bilinear interpolation. The resized features are concatenated along the channel dimension and further aggregated by a convolutional layer to produce a multi-scale localization guidance feature $G^s$. This process allows localization cues from different scales to collaboratively provide coarse-to-fine spatial guidance for the current scale.

Meanwhile, the visual feature $\widetilde{F}_X^s$ is projected into the same channel dimension as $G^s$. The projected visual feature is added to the localization guidance feature $G^s$ and then passed through a $1\times1$ convolution, batch normalization, and sigmoid activation to generate a spatial gating mask $M^s \in[0,1]^{H_s \times W_s}$. The localization-gated feature $\bar{F}_X^s$ is obtained as follows:

\begin{equation}
\label{eq3}
     \bar{F}_X^s=\widetilde{F}_X^s \odot M^s,
\end{equation}where $\odot$ denotes element-wise multiplication. In implementation, the localization maps used in LGAF are detached from the localization prediction branch, so that the gating operation serves as a stable spatial modulation mechanism without directly disturbing the localization prediction branch.

After applying LGAF at four scales, we obtain the localization-gated feature pyramid $\left[\bar{F}_X^1, \bar{F}_X^2, \bar{F}_X^3, \bar{F}_X^4\right]$. These features are subsequently integrated by the segmentation decoder together with high-resolution visual features to produce the final segmentation mask. In this way, LGAF realizes gating-based localization-infused fusion by transforming multi-scale localization predictions into spatial gates.

\subsection{Localization-Constrained Loss Fusion (LCLF)}\label{sec:lclf}
After LGFF and LGAF, the segmentation decoder integrates the localization-gated feature pyramid to predict the final segmentation. Specifically, the decoder follows a top-down six-stage decoding pathway. Starting from the deepest feature $\bar{F}_X^4$, each decoding stage upsamples the higher-level feature and concatenates it with the visual features from LGAF (in the first four stages) or with $F_X^f / F_X^h$ from the visual encoder (in the last two stages), followed by residual convolutional blocks for feature refinement. In addition to the final full-resolution segmentation prediction $\widehat{Y}$, the decoder also produces four intermediate segmentation predictions $\left[\widehat{Y}^1, \widehat{Y}^2, \widehat{Y}^3, \widehat{Y}^4\right]$ from the first four decoding stages. These intermediate predictions are upsampled and aggregated to assist the generation of the final segmentation mask. In this way, the decoder constructs the final prediction by jointly exploiting multi-scale semantics and high-resolution spatial details.

For loss-level localization-infused fusion, Localization-Constrained Loss Fusion (LCLF) jointly optimizes segmentation prediction, localization prediction, and their spatial consistency. The segmentation loss is imposed on the final prediction $\widehat{Y}$ and multi-scale intermediate predictions $\left[\widehat{Y}^1, \widehat{Y}^2, \widehat{Y}^3, \widehat{Y}^4\right]$, where each segmentation loss consists of Dice loss \cite{milletari2016vnet}, focal loss \cite{lin2017focal}, and cross-entropy loss, denoted as $\mathcal{L}_{\mathrm{seg}}$. The localization loss supervises the multi-scale localization predictions $\left[\hat{L}^1, \hat{L}^2, \hat{L}^3, \hat{L}^4\right]$. For each scale $s \in\{4,3,2,1\}$, the corresponding downsampled ground-truth segmentation mask $Y^s$ is divided into patch regions of size $P_s$, and a patch is labeled as positive if the foreground ratio inside the patch exceeds a predefined threshold $\tau$; the resulting patch-wise label map is denoted as $L^s$. Binary cross-entropy loss between $L^s$ and $\hat{L}^s$ is then used to supervise the patch-wise localization prediction, denoted as $\mathcal{L}_{\mathrm{loc}}$. Furthermore, the localization-segmentation consistency loss constrains the predicted segmentation to lie spatially within the corresponding localization map. Each localization map $\hat{L}^s$ is resized to the resolution of the segmentation prediction $\hat{Y}^s$, yielding $\tilde{L}^s$. The consistency loss measures the proportion of predicted foreground pixels that fall inside the predicted localization region:

\begin{equation}
\label{eq4}
    \mathcal{L}_{\mathrm{con}}\left(\hat{Y}^s, \tilde{L}^s\right)=1-\frac{\sum_{i, j} \hat{Y}_{i, j}^s \tilde{L}_{i, j}^s}{\sum_{i, j} \hat{Y}_{i, j}^s}.
\end{equation}

For multi-class segmentation, this loss is computed for each class and averaged over classes and samples. Notably, the localization prediction $\hat{L}^s$ is independently supervised by $\mathcal{L}_{\mathrm{loc}}$ against the patch-wise ground-truth localization labels, which penalizes degenerate all-positive localization maps and thereby discourages $\mathcal{L}_{\mathrm{con}}$ from being trivially minimized, keeping the consistency constraint spatially meaningful.
The overall training objective of LoG is therefore formulated as:

\begin{equation}
\label{eq5}
\begin{split}
    \mathcal{L}_{\mathrm{total}}
    &= \mathcal{L}_{\mathrm{seg}}(Y, \hat{Y})
       + \lambda_{\mathrm{seg}} \sum_{s=1}^4 \mathcal{L}_{\mathrm{seg}}(Y^s, \hat{Y}^s) \\
    &\quad + \lambda_{\mathrm{loc}} \sum_{s=1}^4 \mathcal{L}_{\mathrm{loc}}(L^s, \hat{L}^s)
       + \lambda_{\mathrm{con}} \sum_{s=1}^4 \mathcal{L}_{\mathrm{con}}(\hat{Y}^s, \tilde{L}^s)
\end{split}
\end{equation}where $\lambda_{\mathrm{seg}}$, $\lambda_{\mathrm{loc}}$, and $\lambda_{\mathrm{con}}$ are weighting coefficients. With this objective, LCLF injects localization-oriented semantics into the optimization process by jointly supervising segmentation, localization, and localization-segmentation consistency.

\section{Experimental Setup}
\subsection{Datasets and Preprocessing}

We evaluated LoG on three well-established public benchmark datasets for text-guided medical image segmentation, including QaTa-COV19 \cite{degerli2022osegnet}, MosMedData+ \cite{morozov2020mosmeddata}, and Kvasir-SEG \cite{jha2020kvasirseg}. These datasets cover three representative medical imaging modalities, X-ray, computed tomography (CT), and endoscopic images, and provide paired textual reports for language-guided segmentation.

The QaTa-COV19 dataset contains 9,258 chest X-ray images with ground-truth segmentation masks of COVID-19-infected regions. The official split includes 7,145 training samples and 2,113 testing samples. Following the preprocessing protocol in \cite{ZhongYi2023}, we further divided the official training set into training and validation subsets with a ratio of 80\% and 20\%. The MosMedData+ dataset contains 2,729 CT slices depicting pulmonary infections with corresponding segmentation annotations. Following \cite{LiZihan2024}, we split the dataset into 2,183 training samples, 273 validation samples, and 273 testing samples. For text-guided segmentation, both QaTa-COV19 and MosMedData+ were extended with matched textual reports by Li et al. \cite{LiZihan2024}. Each report consists of three sentences describing the presence of infection, the number of involved regions, and their anatomical locations. These reports provide target-related semantic and spatial cues for lesion segmentation.

The Kvasir-SEG dataset contains 1,000 endoscopic images with ground-truth segmentation masks for gastrointestinal polyps. Following the preprocessing protocol in Poudel et al. \cite{poudel2023medvlsm}, we split the dataset into training, validation, and testing subsets with a ratio of 8:1:1. The paired textual reports describe polyp-related attributes, such as size, number, color, and location, in a free-text format. For all datasets, images and segmentation masks were preprocessed following the corresponding dataset protocols used in previous text-guided segmentation studies \cite{LiZihan2024, ZhongYi2023, LeeGo-Eun2024, ZengQingjie2026, BhardwajRahul2026, XueHengzhi2026}, and the textual reports were tokenized before entering the textual encoder.

\subsection{Implementation Details}
All experiments were implemented in PyTorch and conducted on an NVIDIA RTX 4090 GPU with 24 GB of memory. We trained the models using the Adam optimizer with a piecewise-constant learning-rate schedule. Each model was trained for 100 epochs with a batch size of 32 to ensure sufficient convergence. All images were resized to $224\times224$. During training, data augmentation was applied to improve robustness, including random affine transformations with translation within $\pm10\%$, scaling from 90\% to 110\%, rotation within $\pm 10^{\circ}$, shear within $\pm 10^{\circ}$, and random horizontal flipping with a probability of 0.5. For the loss configuration, the predefined threshold $\tau$ for generating patch-wise localization labels was set to 5\%. The weighting coefficients in the total training objective were empirically determined on the validation sets. Specifically, $\lambda_{\mathrm{seg}}$ and $\lambda_{\mathrm{con}}$ were fixed to 0.1 for all datasets, while $\lambda_{\mathrm{loc}}$ was set to 0.5, 0.4, and 0.2 for the QaTa-COV19, MosMedData+, and Kvasir-SEG datasets, respectively.

\subsection{Experimental Designs}
We compared LoG with three categories of medical image segmentation methods. The first category comprises image-only segmentation methods, including U-Net \cite{Ronneberger2015}, U-Net++ \cite{Zhou2018}, Attention U-Net \cite{oktay2018attention}, Swin-UNet \cite{Cao2023205}, and nnU-Net \cite{IsenseeFabian2020}. The second category comprises vision-language pretraining-based methods, including CLIP \cite{RadfordAlec2021}, GLoRIA \cite{HuangShih-Cheng2022}, BiomedCLIP \cite{zhang2025multimodal}, and MedSAM \cite{ma2024medsam}, which exploit image-text or foundation-model pretraining for downstream segmentation. The third category comprises text-guided medical image segmentation methods, including LViT \cite{LiZihan2024}, Ariadne \cite{ZhongYi2023}, MAdaptor \cite{zhang2024madapter}, FMISeg \cite{yu2025fmiseg}, EviVLM \cite{pan2026evivlm}, RecLMIS \cite{huang2025reclmis}, TeViA \cite{ZengQingjie2026}, ViTexNet \cite{BhardwajRahul2026}, and PCSA-Seg \cite{XueHengzhi2026}. We used the official results reported in the corresponding papers when available; otherwise, we re-implemented the methods following their original papers or code. CLIP and GLoRIA were not evaluated on the endoscopic polyp dataset (Kvasir-SEG), as they were not pretrained on this domain. The Dice Similarity Coefficient (DSC) and mean Intersection over Union (mIoU) were adopted as evaluation metrics.

We conducted ablation studies to evaluate the effectiveness of the proposed localization-infused fusion designs. First, component-wise ablation was performed by progressively adding the three fusion components (LGFF, LGAF, LCLF) to a baseline model. The baseline directly fuses textual and visual features in the skip connections between the visual encoder and the segmentation decoder without applying localization-oriented semantic extraction and fusion. Second, component-specific ablation was conducted to analyze the design choices of each fusion component. All ablation analyses were conducted on the three datasets using a unified protocol, where all settings were kept identical except for the ablated component.

We further performed visualization analysis. First, qualitative comparisons were conducted between LoG and representative comparison methods to show the improvement in ROI delineation, especially in challenging cases with ambiguous boundaries or subtle target regions. Then, we visualized the predicted localization maps, the spatial gating masks generated by LGAF, and the final segmentation predictions to examine their spatial consistency. For visualization, the multi-scale localization maps and spatial gating masks were first upsampled to the original image size and then averaged across four scales, producing aggregated localization and gating maps for visual comparison with segmentation masks.

In addition, we conducted sensitivity analyses to explore how the granularity of patch-wise localization prediction affects segmentation performance by evaluating different settings of localization patch size $P_s$. Finally, we evaluated LoG's performance under textual input truncation, where the input report is progressively truncated to different token lengths.

\section{Results and Discussion}
\subsection{Comparison with Existing Methods}

Table~\ref{tab1} reports the quantitative comparison between LoG and existing methods on QaTa-COV19, MosMedData+, and Kvasir-SEG. Overall, text-guided segmentation methods achieved better performance than image-only and pretraining-based methods across the three datasets, demonstrating the effectiveness of incorporating textual guidance into medical image segmentation. Compared with image-only methods, clinical reports provide crucial semantic cues about target identity, appearance, and location, which are particularly useful for segmenting ambiguous or low-contrast ROIs.

\begin{table*}[t]
\caption{Quantitative Comparison Between LoG and Existing Methods for Medical Image Segmentation}
\label{tab1}
\centering
\footnotesize

\begingroup
\setlength{\tabcolsep}{6pt}
\setlength{\arrayrulewidth}{0.30pt}
\setlength{\heavyrulewidth}{0.90pt}
\setlength{\lightrulewidth}{0.40pt}
\setlength{\cmidrulewidth}{0.40pt}
\renewcommand{\arraystretch}{1.08}

\setbox0=\hbox{%
\begin{tabular}{@{}p{3.50cm}|c|c|c|c|c|c|c|c@{}}

\toprule

\multicolumn{1}{c|}{%
  \raisebox{-1.75ex}[0pt][0pt]{\textbf{Method}}%
}
& \multicolumn{2}{c|}{\textbf{QaTa-COV19}}
& \multicolumn{2}{c|}{\textbf{MosMedData+}}
& \multicolumn{2}{c|}{\textbf{Kvasir-SEG}}
& \multicolumn{2}{c}{\textbf{Average}}
\\[-0.15ex]

\cmidrule(lr){2-3}
\cmidrule(lr){4-5}
\cmidrule(lr){6-7}
\cmidrule(l){8-9}

\multicolumn{1}{c|}{}
& \multicolumn{1}{c|}{\textbf{Dice (\%)}}
& \multicolumn{1}{c|}{\textbf{mIoU (\%)}}
& \multicolumn{1}{c|}{\textbf{Dice (\%)}}
& \multicolumn{1}{c|}{\textbf{mIoU (\%)}}
& \multicolumn{1}{c|}{\textbf{Dice (\%)}}
& \multicolumn{1}{c|}{\textbf{mIoU (\%)}}
& \multicolumn{1}{c|}{\textbf{Dice (\%)}}
& \multicolumn{1}{c}{\textbf{mIoU (\%)}}
\\

\midrule

\multicolumn{9}{@{}l@{}}{%
  \rule{0pt}{2.30ex}%
  \hspace*{0.20em}%
  \textbf{\textit{Image-only methods}}%
}
\\[-0.10ex]

\hspace*{0.75em}U-Net
& 82.99 & 70.92 & 64.60 & 50.73 & 82.09 & 69.62 & 76.56 & 63.76
\\

\hspace*{0.75em}U-Net++
& 83.69 & 71.96 & 71.75 & 58.39 & 82.06 & 69.58 & 79.17 & 66.64
\\

\hspace*{0.75em}Attention U-Net
& 82.40 & 70.06 & 66.34 & 52.82 & 85.18 & 74.19 & 77.97 & 65.69
\\

\hspace*{0.75em}Swin-UNet
& 83.60 & 72.40 & 66.90 & 53.10 & 82.41 & 72.38 & 77.64 & 65.96
\\

\hspace*{0.75em}nnU-Net
& 80.42 & 70.81 & 72.59 & 60.36 & 85.18 & 75.48 & 79.40 & 68.88
\\

\midrule

\multicolumn{9}{@{}l@{}}{%
  \rule{0pt}{2.30ex}%
  \hspace*{0.20em}%
  \textbf{\textit{Pretraining-based methods}}%
}
\\[-0.10ex]

\hspace*{0.75em}CLIP
& 79.81 & 70.66 & 71.97 & 59.64 & -- & -- & 75.89$^{\dagger}$ & 65.15$^{\dagger}$
\\

\hspace*{0.75em}GLoRIA
& 79.94 & 70.68 & 72.42 & 60.18 & -- & -- & 76.18$^{\dagger}$ & 65.43$^{\dagger}$
\\

\hspace*{0.75em}BiomedCLIP
& 87.88 & 78.38 & 66.51 & 50.52 & 85.66 & 77.91 & 80.02 & 68.94
\\

\hspace*{0.75em}MedSAM
& 85.78 & 78.62 & 75.24 & 62.43 & 89.12 & 82.45 & 83.38 & 74.50
\\

\midrule

\multicolumn{9}{@{}l@{}}{%
  \rule{0pt}{2.30ex}%
  \hspace*{0.20em}%
  \textbf{\textit{Text-guided methods}}%
}
\\[-0.10ex]

\hspace*{0.75em}LViT
& 84.92 & 73.79 & 74.57 & 61.33 & 87.03 & 77.04 & 82.17 & 70.72
\\

\hspace*{0.75em}Ariadne
& 89.78 & 81.45 & 77.33 & 63.04 & 89.32 & 80.70 & 85.48 & 75.06
\\

\hspace*{0.75em}MAdaptor
& 90.22 & 82.16 & 78.62 & 64.78 & 91.58 & 84.47 & 86.81 & 77.14
\\

\hspace*{0.75em}FMISeg
& 91.21
& 83.84
& \underline{79.30}
& 65.71
& 91.12
& 83.52
& 87.21
& 77.69
\\

\hspace*{0.75em}EviVLM
& 85.79
& 77.34
& 77.64
& \underline{65.81}
& 89.63
& 82.45
& 84.35
& 75.20
\\

\hspace*{0.75em}RecLMIS
& 85.22 & 77.00 & 77.48 & 65.07 & 85.75 & 78.76 & 82.82 & 73.61
\\

\hspace*{0.75em}TeViA
& 91.06 & 83.60 & 78.49 & 64.59 & 90.92 & 83.77 & 86.82 & 77.32
\\

\hspace*{0.75em}ViTexNet
& 90.76 & 83.25 & 78.19 & 64.04 & 91.44 & 84.62 & 86.80 & 77.30
\\

\hspace*{0.75em}PCSA-Seg
& \underline{91.43}
& \underline{84.21}
& 78.60
& 64.74
& \underline{92.24}
& \underline{85.15}
& \underline{87.42}
& \underline{78.03}
\\

\hspace*{0.75em}LoG (Ours)
& \textbf{91.59}
& \textbf{84.48}
& \textbf{80.71}
& \textbf{67.66}
& \textbf{94.59}
& \textbf{89.73}
& \textbf{88.96}
& \textbf{80.62}
\\

\bottomrule

\end{tabular}%
}

\makebox[\linewidth][c]{\copy0}
\par
\vspace{1.5pt}

\makebox[\linewidth][c]{
  \parbox{\wd0}{
    \raggedright
    \footnotesize
    \textbf{Bold} and \underline{underline} denote the best and
    second-best results in each column. 
    $\dagger$: Averaged over QaTa-COV19 and MosMedData+ only.
  }
}

\endgroup
\end{table*}

Among all compared methods, LoG achieved the best performance across the three datasets and consistently outperformed existing text-guided segmentation methods. PCSA-Seg achieved the overall second-best performance with a sophisticated image-text feature fusion strategy; nevertheless, it did not surpass LoG. The advantage of LoG is particularly evident on Kvasir-SEG, where foreground polyps often exhibit high visual similarity to surrounding background tissues. In such challenging cases, textual descriptions of ROI location and attributes are particularly useful for distinguishing target regions, and LoG more effectively transforms such textual cues into localization-oriented segmentation guidance.

The superior performance of LoG is attributed to the proposed localization-infused vision-language fusion framework. Unlike existing text-guided methods that mainly rely on holistic textual features and feature-level image-text fusion, LoG explicitly extracts localization-oriented semantics and injects them into segmentation through LGFF, LGAF, and LCLF. These results indicate that transforming textual information into explicit localization cues for multi-level information fusion is beneficial for accurate ROI delineation. The effectiveness of each proposed component is further analyzed in the following ablation analyses.

\subsection{Ablation Analysis}
Table~\ref{tab2} presents the ablation analysis in which LGFF, LGAF, and LCLF are progressively incorporated into the baseline model. Incorporating LGFF contributed to the largest performance gain, improving the average DSC by 2.17\% across the three datasets. This improvement demonstrates that explicitly extracting localization-oriented semantics to guide feature-level fusion is highly effective for text-guided segmentation. Further adding LGAF yielded an additional 0.95\% average DSC improvement, indicating that explicit localization predictions can provide useful spatial gating cues to reinforce target-relevant regions. Finally, incorporating LCLF further improved the average DSC by 0.69\%, showing that localization cues can also benefit segmentation through loss-level spatial consistency constraints. These results verify the complementarity of the proposed three-level fusion strategy: beyond implicit feature-space fusion through LGFF, explicit localization cues can be further injected into segmentation via gating-based modulation and optimization-level constraints, leading to consistent performance gains across all datasets.

\begin{table}[t]
\caption{DSC (\%) Results of Progressively Adding Each Component}
\label{tab2}
\centering
\scriptsize

\begingroup
\setlength{\tabcolsep}{3.2pt}
\setlength{\arrayrulewidth}{0.30pt}
\setlength{\heavyrulewidth}{0.90pt}
\setlength{\lightrulewidth}{0.40pt}
\renewcommand{\arraystretch}{1.14}

\setbox0=\hbox{%
\begin{tabular}{@{}ccc|ccc|c@{}}

\toprule

\textbf{LGFF}
& \textbf{LGAF}
& \textbf{LCLF}
& \textbf{QaTa-COV19}
& \textbf{MosMedData+}
& \textbf{Kvasir-SEG}
& \textbf{Average}
\\

\midrule

$\times$
& $\times$
& $\times$
& 88.25
& 76.84
& 90.36
& 85.15
\\

$\surd$
& $\times$
& $\times$
& 90.28
& 78.86
& 92.82
& 87.32
\\

$\surd$
& $\surd$
& $\times$
& \underline{90.91}
& \underline{79.94}
& \underline{93.96}
& \underline{88.27}
\\

$\surd$
& $\surd$
& $\surd$
& \textbf{91.59}
& \textbf{80.71}
& \textbf{94.59}
& \textbf{88.96}
\\

\bottomrule

\end{tabular}%
}

\makebox[\columnwidth][c]{\copy0}
\par
\vspace{1.5pt}

\makebox[\columnwidth][c]{%
  \parbox{\wd0}{%
    \raggedright
    \scriptsize
    \textbf{Bold} and \underline{underline} denote the best and
    second-best results in each column.%
  }%
}

\endgroup
\end{table}

Table~\ref{tab3} presents the ablation analysis on LGFF. Removing the progressive update of textual feature $F_T$ caused an average DSC drop of 0.29\%, indicating that updating textual representations with scale-specific visual contexts is beneficial for coarse-to-fine vision-language interaction. When feature fusion was removed, although localization prediction was still performed and the subsequent LGAF and LCLF remained unchanged, the average DSC decreased by 1.10\%. This suggests that, although LGAF and LCLF enable explicit localization-oriented semantic guidance via gating-based modulation and loss-level constraints, directly injecting semantic cues via cross-attention in the feature space is still beneficial. The most significant degradation occurred when the localization prediction task was removed, leading to an average DSC drop of 3.44\%. In this setting, the localization prediction head was discarded, causing LGFF, LGAF, and LCLF to lose their localization-oriented basis, and textual features were directly fused with visual features without explicit localization guidance. This result confirms that explicit localization prediction is the core foundation of LoG, enabling textual semantics to be transformed into target-oriented cues for multi-level fusion.

\begin{table}[t]
\caption{DSC (\%) Results of Ablation Analysis on LGFF}
\label{tab3}
\centering
\scriptsize

\begingroup
\setlength{\tabcolsep}{3.2pt}
\setlength{\arrayrulewidth}{0.30pt}
\setlength{\heavyrulewidth}{0.90pt}
\setlength{\lightrulewidth}{0.40pt}
\renewcommand{\arraystretch}{1.14}

\setbox0=\hbox{%
\begin{tabular}{@{}l|ccc|c@{}}

\toprule

\multicolumn{1}{c|}{\textbf{Method}}
& \textbf{QaTa-COV19}
& \textbf{MosMedData+}
& \textbf{Kvasir-SEG}
& \textbf{Average}
\\

\midrule

Full model (Ours)
& \textbf{91.59}
& \textbf{80.71}
& \textbf{94.59}
& \textbf{88.96}
\\

Without updating $F_T$
& \underline{91.34}
& \underline{80.44}
& \underline{94.22}
& \underline{88.67}
\\

Without feature fusion
& 90.75
& 79.64
& 93.19
& 87.86
\\

Without localization$^{\dagger}$
& 88.84
& 77.21
& 90.51
& 85.52
\\

\bottomrule

\end{tabular}%
}

\makebox[\columnwidth][c]{\copy0}
\par
\vspace{1.5pt}

\makebox[\columnwidth][c]{%
  \parbox{\wd0}{%
    \raggedright
    \scriptsize
    \textbf{Bold} and \underline{underline} denote the best and
    second-best results in each column. \\$^{\dagger}$ LGAF and LCLF are
    removed as well.%
  }%
}

\endgroup
\end{table}

Table~\ref{tab4} presents the ablation analysis on LGAF. When relying merely on single-scale localization prediction $\hat{L}^s$ to generate $G^s$, the average DSC dropped by 0.41\%, indicating that using only the localization prediction from the corresponding scale is suboptimal. This suggests that localization cues from different scales are complementary: coarse-scale predictions provide global target awareness, while fine-scale predictions offer more detailed spatial guidance. When removing localization information, LGAF degraded into a basic spatial attention mechanism driven only by visual features, and the average DSC decreased by 1.00\%, which is very close to the 1.05\% drop caused by directly removing LGAF. This indicates that a generic spatial attention mechanism contributes little without explicit localization guidance, yielding performance comparable to using no gating operation at all. The performance drop caused by removing LGAF validates the individual effectiveness of LGAF.

\begin{table}[t]
\caption{DSC (\%) Results of Ablation Analysis on LGAF}
\label{tab4}
\centering
\scriptsize

\begingroup
\setlength{\tabcolsep}{3.2pt}
\setlength{\arrayrulewidth}{0.30pt}
\setlength{\heavyrulewidth}{0.90pt}
\setlength{\lightrulewidth}{0.40pt}
\renewcommand{\arraystretch}{1.14}

\setbox0=\hbox{%
\begin{tabular}{@{}l|ccc|c@{}}

\toprule

\multicolumn{1}{c|}{\textbf{Method}}
& \textbf{QaTa-COV19}
& \textbf{MosMedData+}
& \textbf{Kvasir-SEG}
& \textbf{Average}
\\

\midrule

Full model (Ours)
& \textbf{91.59}
& \textbf{80.71}
& \textbf{94.59}
& \textbf{88.96}
\\

With single-scale LocInfo
& \underline{91.24}
& \underline{80.26}
& \underline{94.15}
& \underline{88.55}
\\

Without LocInfo
& 90.81
& 79.58
& 93.50
& 87.96
\\

Without LGAF
& 90.76
& 79.54
& 93.42
& 87.91
\\

\bottomrule

\end{tabular}%
}

\makebox[\columnwidth][c]{\copy0}
\par
\vspace{1.5pt}

\makebox[\columnwidth][c]{%
  \parbox{\wd0}{%
    \raggedright
    \scriptsize
    \textbf{Bold} and \underline{underline} denote the best and
    second-best results in each column. LocInfo denotes localization
    information ($\hat{L}^{s}$).%
  }%
}

\endgroup
\end{table}

Table~\ref{tab5} presents the ablation analysis on LCLF, in which the localization-segmentation consistency loss was progressively removed from different scales. A consistent performance degradation is observed as the consistency constraint is removed from more scales. Removing $\mathcal{L}_{\mathrm{con}}$ at scale $s=1$ caused an average DSC drop of 0.25\%, which is the largest incremental degradation among the four scales. This indicates that the highest-resolution localization prediction provides finer-grained spatial cues and plays a crucial role in constraining ROI delineation. As $\mathcal{L}_{\mathrm{con}}$ is progressively removed from scales $s \in\{1,2\}$ and $s \in\{1,2,3\}$, the average DSC further decreased by 0.19\% and 0.14\%, respectively. When the consistency loss is removed from all four scales, the average DSC drops by 0.69\% compared with the full model, confirming the overall effectiveness of LCLF.

\begin{table}[t]
\caption{DSC (\%) Results of Ablation Analysis on LCLF}
\label{tab5}
\centering
\scriptsize

\begingroup
\setlength{\tabcolsep}{3.2pt}
\setlength{\arrayrulewidth}{0.30pt}
\setlength{\heavyrulewidth}{0.90pt}
\setlength{\lightrulewidth}{0.40pt}
\renewcommand{\arraystretch}{1.10}

\setbox0=\hbox{%
\begin{tabular}{@{}l|ccc|c@{}}

\toprule

\multicolumn{1}{c|}{\textbf{Method}}
& \textbf{QaTa-COV19}
& \textbf{MosMedData+}
& \textbf{Kvasir-SEG}
& \textbf{Average}
\\

\midrule

Full model (Ours)
& \textbf{91.59}
& \textbf{80.71}
& \textbf{94.59}
& \textbf{88.96}
\\

\shortstack[l]{w/o $\mathcal{L}_{\mathrm{con}}$,
$s \in \{1\}$}& \underline{91.35}
& \underline{80.45}
& \underline{94.32}
& \underline{88.71}
\\

\shortstack[l]{w/o $\mathcal{L}_{\mathrm{con}}$,
$s \in \{1,2\}$}& 91.17
& 80.24
& 94.15
& 88.52
\\

\shortstack[l]{w/o $\mathcal{L}_{\mathrm{con}}$,
$s \in \{1,2,3\}$}& 91.02
& 80.07
& 94.04
& 88.38
\\

\shortstack[l]{w/o $\mathcal{L}_{\mathrm{con}}$,
$s \in \{1,2,3,4\}$}& 90.91
& 79.94
& 93.96
& 88.27
\\

\bottomrule

\end{tabular}%
}

\makebox[\columnwidth][c]{\copy0}
\par
\vspace{1.5pt}

\makebox[\columnwidth][c]{%
  \parbox{\wd0}{%
    \raggedright
    \scriptsize
    \textbf{Bold} and \underline{underline} denote the best and
    second-best results in each column.%
  }%
}

\endgroup
\end{table}

\subsection{Visualization Analysis}
\begin{figure*}[t]
    \centering
    \includegraphics[
        width=0.9\textwidth,
        keepaspectratio
    ]{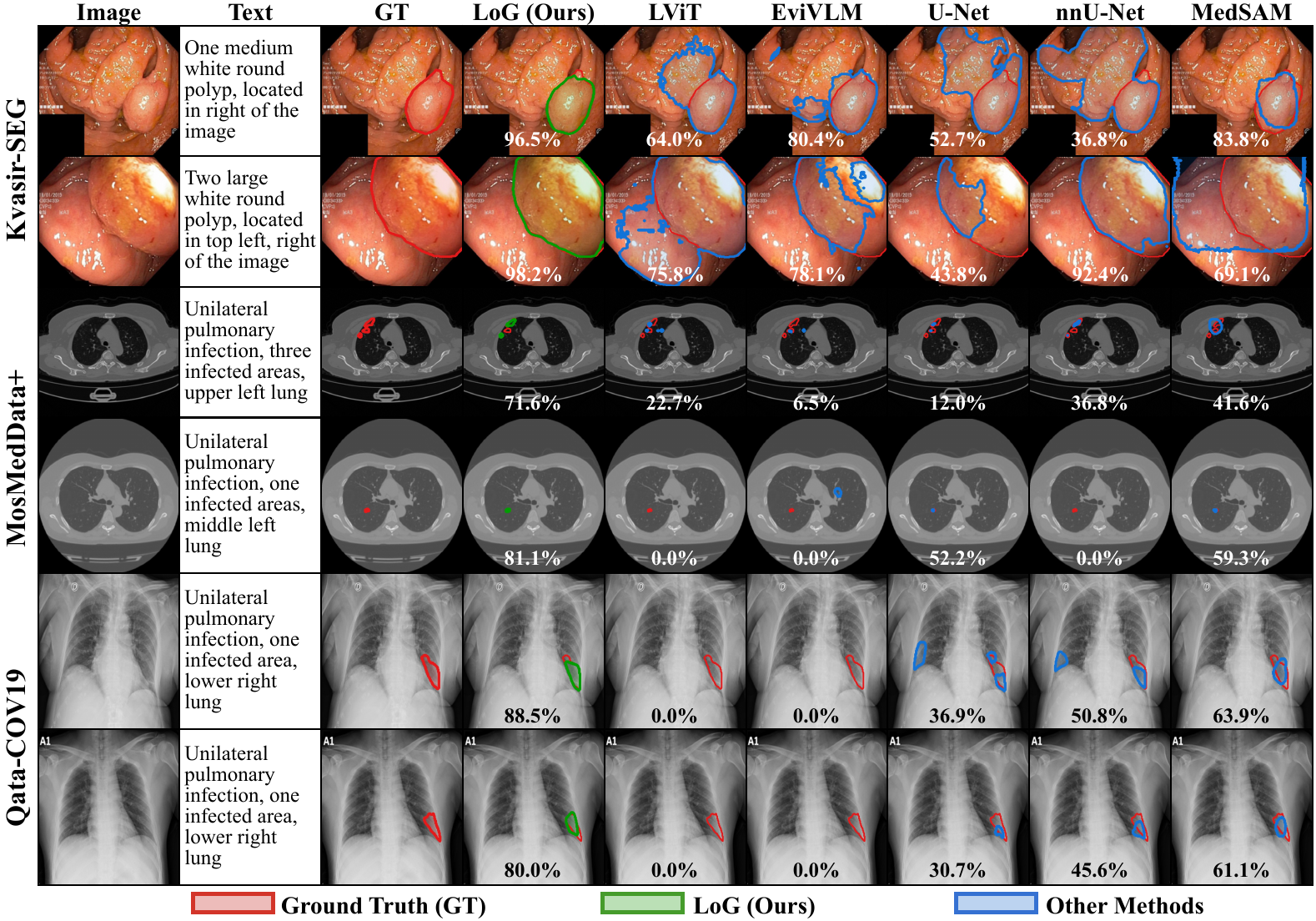}
    \caption{Qualitative comparison with representative medical image
    segmentation methods on Kvasir-SEG (rows 1--2), MosMedData+
    (rows 3--4), and QaTa-COV19 (rows 5--6). The DSC score is shown
    below each prediction. Red, green, and blue contours denote the
    ground truth, predictions of LoG, and predictions of the compared
    methods, respectively.}
    \label{fig5}
\end{figure*}

Fig.~\ref{fig5} provides qualitative comparisons between LoG and representative comparison methods on Kvasir-SEG, MosMedData+, and QaTa-COV19. Overall, LoG produces segmentation masks that are more spatially consistent with the ground truth and better aligned with the textual descriptions. On Kvasir-SEG, where the polyp regions often exhibit similar color and texture to surrounding tissues, LoG accurately delineated the target polyps and preserved clearer boundaries, while other methods tended to generate coarser or incomplete predictions. On MosMedData+ and QaTa-COV19, the target infection regions are often small, low-contrast, or visually ambiguous. In these cases, several comparison methods missed the target regions or produced obvious location drift, whereas LoG maintained more reliable localization and delineation. These visual results further support the quantitative findings in Table~\ref{tab1}, showing that LoG can more effectively exploit textual semantics and transform them into localization-oriented segmentation guidance, especially under challenging conditions where visual cues alone are insufficient.

\begin{figure*}[t]
    \raggedright
    \includegraphics[
        width=0.96\textwidth,
        keepaspectratio
    ]{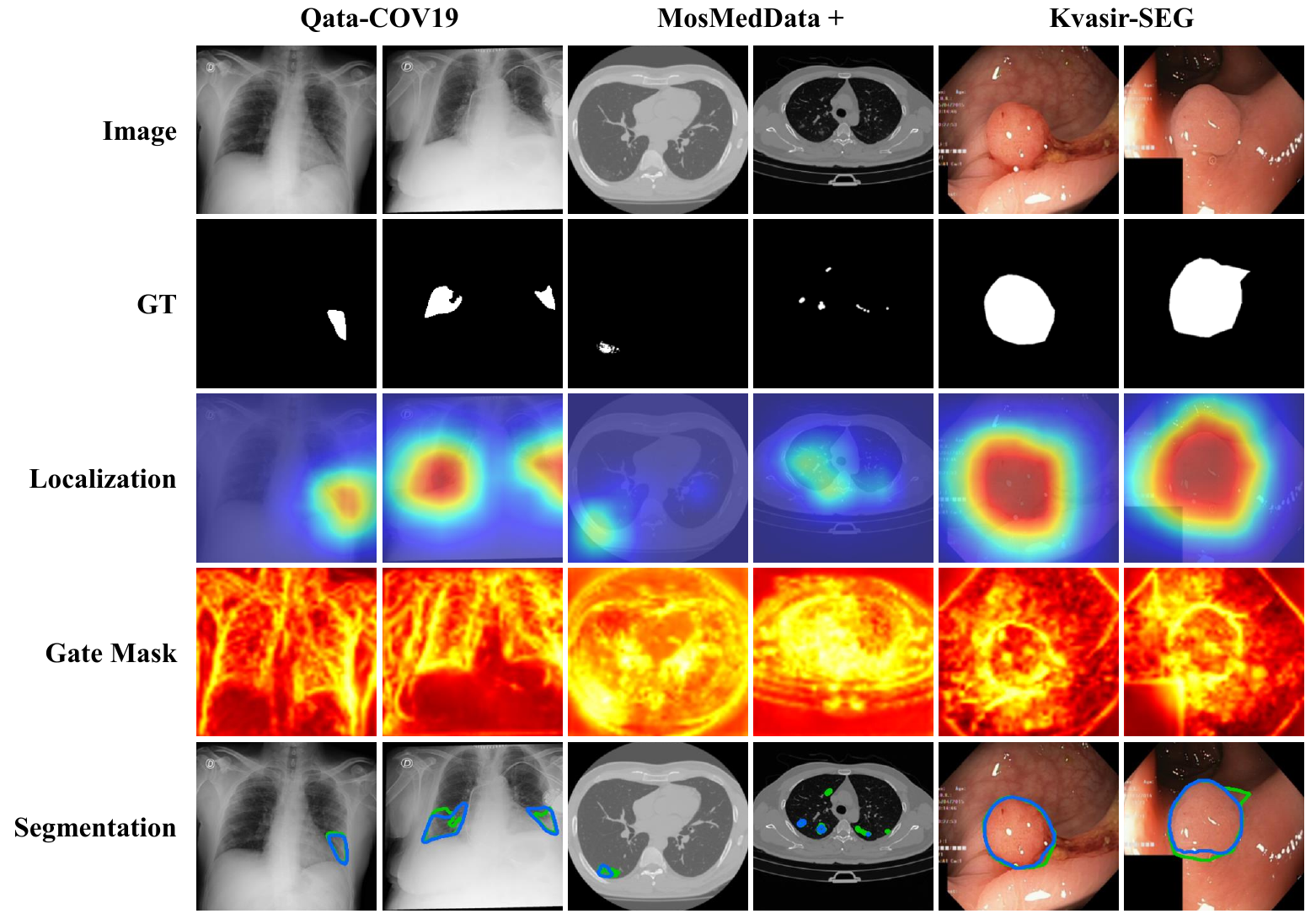}
    \caption{Visualization of the input image, ground truth (GT),
    localization map, LGAF gating mask, and segmentation prediction
    from left to right. Green and blue contours denote the ground truth
    and segmentation prediction, respectively.}
    \label{fig6}
\end{figure*}

Fig.~\ref{fig6} further visualizes the internal localization behavior of LoG, including the input image, ground truth segmentation mask, predicted localization map, LGAF gating mask, and final segmentation result. Across the three datasets, the predicted localization maps consistently highlight regions that correspond well to the ground-truth ROIs, indicating that the localization-oriented semantic extraction module can successfully transform textual guidance into spatially meaningful localization cues. The LGAF gating masks also show strong spatial consistency with the localization maps, suggesting that the predicted localization information is effectively redirected into the segmentation pathway to enhance target-relevant regions. As a result, the final segmentation predictions are well aligned with both the ground truth and the localization responses. These observations demonstrate that LoG does not merely use textual features as global semantic embeddings; instead, it explicitly converts textual cues into localization-oriented semantic guidance, leading to spatially consistent and accurate segmentation.

\subsection{Sensitivity Analysis}
Table~\ref{tab6} presents the sensitivity analysis of the localization patch size $P_s$. Overall, LoG is not highly sensitive to the selection of $P_s$, as all evaluated settings achieved competitive performance. This indicates that the proposed localization-oriented semantic extraction is robust to different localization granularities.

When larger patch sizes are used, the localization prediction becomes coarse, where each localization prediction corresponds to a larger image region. This may weaken the spatial precision of localization cues and lead to a slight performance decrease. In contrast, reducing $P_s$ produces finer-grained localization predictions, which can provide more detailed spatial guidance for segmentation and therefore brings performance improvements. However, excessively reducing $P_s$ only led to marginal gains while increasing computational cost due to the larger number of localization tokens. Therefore, we select the second-row setting in Table~\ref{tab6} as the default configuration, which provides a favorable balance between segmentation performance and computational efficiency.

\begin{table}[t]
\caption{DSC (\%) Results of Sensitivity Analysis on $P_s$}
\label{tab6}
\centering
\scriptsize

\begingroup
\setlength{\tabcolsep}{3.2pt}
\setlength{\arrayrulewidth}{0.30pt}
\setlength{\heavyrulewidth}{0.90pt}
\setlength{\lightrulewidth}{0.40pt}
\renewcommand{\arraystretch}{1.14}

\setbox0=\hbox{%
\begin{tabular}{@{}cccc|ccc|c@{}}

\toprule

{\bfseries\boldmath $P_1$}
& {\bfseries\boldmath $P_2$}
& {\bfseries\boldmath $P_3$}
& {\bfseries\boldmath $P_4$}
& \textbf{QaTa-COV19}
& \textbf{MosMedData+}
& \textbf{Kvasir-SEG}
& \textbf{Average}
\\

\midrule

$7$
& $6$
& $5$
& $4$
& 91.24
& 80.25
& 93.98
& 88.49
\\

{\bfseries\boldmath $6$}
& {\bfseries\boldmath $5$}
& {\bfseries\boldmath $4$}
& {\bfseries\boldmath $3$}
& \textbf{91.59}
& \textbf{80.71}
& \textbf{94.59}
& \textbf{88.96}
\\

$5$
& $4$
& $3$
& $2$
& 91.61
& 80.74
& 94.65
& 89.00
\\

$4$
& $3$
& $2$
& $1$
& 91.62
& 80.75
& 94.67
& 89.01
\\

\bottomrule

\end{tabular}%
}

\makebox[\columnwidth][c]{\copy0}
\par
\vspace{1.5pt}

\makebox[\columnwidth][c]{%
  \parbox{\wd0}{%
    \raggedright
    \scriptsize
    \textbf{Bold} denotes the default settings used in the experiments.%
  }%
}

\endgroup
\end{table}

\subsection{Textual Input Analysis}
Table~\ref{tab7} reports the DSC results under textual input truncation. In our implementation, the textual encoder uses \texttt{max\_length=24} with \texttt{padding='max\_length'} by default, while the average report length is approximately 11.5--12.5 tokens in the evaluation datasets. Specifically, the first four tokens often cover only diagnostic phrases, tokens 5--8 contain key spatial localization descriptions, tokens 9--12 contain supplemental morphology, and later tokens are largely redundant. Therefore, truncating the input to 16 or 12 tokens mainly removed redundant padding and thus slightly improved the average DSC. When the textual input is further truncated to 8 and 4 tokens, the average DSC drops by 0.54\% and 2.35\%, respectively. This degradation is reasonable because excessive truncation may remove key location cues required for accurate localization-oriented fusion.

When no text is provided, the average DSC decreases by 2.97\% compared with the original-text setting, clearly demonstrating the contribution of textual guidance. Without textual input, LoG can only rely on visual information to perform localization prediction as semantic guidance, leading to obvious performance degradation. Nevertheless, the no-text variant still achieves competitive performance compared with many existing text-guided segmentation methods in Table~\ref{tab1}. This suggests that the proposed localization-oriented strategy is useful even without textual reports, because explicitly encouraging the model to predict target location can still help extract spatial cues from visual features. In fact, the no-text variant (85.99\% average DSC) remains higher than the ``Without localization'' variant in Table~\ref{tab3} (85.52\%), which discards the entire localization-oriented design. This indicates that the localization-oriented multi-task formulation itself is a primary source of the improvement, while paired textual reports act as complementary guidance that further sharpens localization by injecting explicit spatial semantics, together yielding the best performance.

\begin{table}[t]
\caption{DSC (\%) Results Under Textual Input Truncation}
\label{tab7}
\centering
\scriptsize

\begingroup
\setlength{\tabcolsep}{3.2pt}
\setlength{\arrayrulewidth}{0.30pt}
\setlength{\heavyrulewidth}{0.90pt}
\setlength{\lightrulewidth}{0.40pt}
\renewcommand{\arraystretch}{1.14}

\setbox0=\hbox{%
\begin{tabular}{@{}l|ccc|c@{}}

\toprule

\multicolumn{1}{c|}{\textbf{Method}}
& \textbf{QaTa-COV19}
& \textbf{MosMedData+}
& \textbf{Kvasir-SEG}
& \textbf{Average}
\\

\midrule

Original text (Ours)
& 91.59
& 80.71
& 94.59
& 88.96
\\

Truncated to 16 tokens
& \underline{91.59}
& \underline{80.72}
& \underline{94.60}
& \underline{88.97}
\\

Truncated to 12 tokens
& \textbf{91.60}
& \textbf{80.73}
& \textbf{94.62}
& \textbf{88.98}
\\

Truncated to 8 tokens
& 91.23
& 80.15
& 93.89
& 88.42
\\

Truncated to 4 tokens
& 89.78
& 78.92
& 91.12
& 86.61
\\

No text
& 89.28
& 78.34
& 90.35
& 85.99
\\

\bottomrule

\end{tabular}%
}

\makebox[\columnwidth][c]{\copy0}
\par
\vspace{1.5pt}

\makebox[\columnwidth][c]{%
  \parbox{\wd0}{%
    \raggedright
    \scriptsize
    \textbf{Bold} and \underline{underline} denote the best and
    second-best results in each column.%
  }%
}

\endgroup
\end{table}

\section{Conclusion and Limitations}
In this study, we have outlined LoG, a localization-infused vision-language fusion framework for text-guided medical image segmentation. Unlike existing methods that typically encode textual reports as holistic semantic embeddings and fuse them with visual features mainly at the feature level, LoG reformulates text-guided segmentation as a localization-oriented semantic extraction and fusion problem. By jointly performing multi-scale localization, LoG explicitly extracts target-relevant vision-language semantics and represents them in both implicit feature and explicit localization-prediction spaces. Based on these semantics, three localization-infused fusion strategies were introduced, including Localization-Guided Feature Fusion (LGFF), Localization-Gated Attention Fusion (LGAF), and Localization-Constrained Loss Fusion (LCLF), which inject localization cues into segmentation through feature-level fusion, gating-based modulation, and loss-level spatial constraints, respectively. Extensive experiments on QaTa-COV19, MosMedData+, and Kvasir-SEG demonstrated that LoG consistently outperformed existing image-only, pretraining-based, and text-guided segmentation methods. Ablation, visualization, sensitivity, and textual input analyses further verified the effectiveness of multi-scale localization prediction and multi-level localization-infused information fusion in the context of text-guided medical image segmentation.

Our study still has several limitations. First, LoG relies on paired textual reports as auxiliary input, and its performance may be affected when textual information is incomplete, unavailable, or inconsistent with the image content. Second, we validated LoG on structured reports. In real clinical scenarios, reports are more diverse and may be noisy, which may introduce additional challenges for robust textual semantic extraction. Third, the current experiments were conducted on 2D medical images from three benchmark datasets, while extension to more anatomical regions, imaging modalities, and 3D volumetric segmentation tasks remains to be investigated. In future work, we will explore more robust language understanding strategies, such as LLM-based report parsing and region-text grounding, to handle diverse clinical reports and further improve localization-oriented vision-language fusion in broader medical image segmentation scenarios.

\bibliographystyle{IEEEtran}
\bibliography{reference}

\end{document}